\documentclass[review]{elsarticle}

\usepackage{times}
\usepackage{graphicx}
\usepackage{latexsym}
\usepackage{amsmath,amsfonts,amssymb,dsfont,url}
\usepackage{subfig}
\usepackage[utf8]{inputenc}
\usepackage[T1]{fontenc}

\usepackage{hyperref}

\journal{Journal of \LaTeX\ Templates}

\usepackage{tikz}
\usetikzlibrary{arrows, decorations.markings, positioning}
\usetikzlibrary{arrows.meta,shapes.arrows}

\def\R{\mathbb{R}}
\def\M{\mathcal{M}}
\def\C{\mathcal{C}}
\def\LL{\mathcal{L}}
\def\X{\mathcal{X}}
\def\U{\mathcal{U}}
\def\1{\mathds{1}}
\def\our{S$^3$C$^2$}
\def\SLN{LabNet}

\def\CN{CluNet}

\def\sln{\mathrm{L^{\tiny contr}}}

\def\mse{\mathrm{L^{\tiny recon}}}
\def\cn{\mathrm{L^{\tiny miscl}}}

\begin{document}

\begin{frontmatter}

\title{A Classification-Based Approach to Semi-Supervised Clustering with Pairwise Constraints}

\author[mymainaddress]{Marek \'Smieja\corref{mycorrespondingauthor}}
\cortext[mycorrespondingauthor]{Corresponding author}
\ead[url]{marek.smieja@uj.edu.pl}

\author[mymainaddress]{\L{}ukasz Struski}
\ead[url]{lukasz.struski@uj.edu.pl}

\author[mysecondaryaddress]{Mário A. T. Figueiredo}
\ead[url]{mario.figueiredo@tecnico.ulisboa.pt}

\address[mymainaddress]{Faculty of Mathematics and Computer Science, Jagiellonian University, Krak\'ow, Poland}

\address[mysecondaryaddress]{Instituto de Telecomunicações,  Instituto Superior Técnico, Universidade de Lisboa, Lisbon, Portugal}

\begin{abstract}
In this paper, we introduce a neural network framework for semi-supervised clustering (SSC) with pairwise (must-link or cannot-link) constraints. In contrast to existing approaches, we decompose SSC into two simpler classification tasks/stages: the first stage uses a pair of Siamese neural networks to label the unlabeled pairs of points as must-link or cannot-link; the second stage uses the fully pairwise-labeled dataset produced by the first stage in a supervised neural-network-based clustering method. The proposed approach, \our{} ({\bf S}emi-{\bf S}upervised \textbf{S}iamese \textbf{C}lassifiers for \textbf{C}lustering), is motivated by the observation that binary classification (such as assigning pairwise relations) is usually easier than multi-class clustering with partial supervision. On the other hand, being classification-based, our method solves only well-defined classification problems, rather than less well specified clustering tasks. Extensive experiments on various datasets demonstrate the high performance of the proposed method.
\end{abstract}


\end{frontmatter}

\section{Introduction}

Clustering is an important unsupervised learning tool often used to analyze the structure of complex high-dimensional data. Without any additional information about the underlying class/cluster structure, clustering results may contradict our prior knowledge or assumptions about the data being analyzed. \textit{Semi-supervised clustering} (SSC) methods tackle this issue by leveraging partial prior information about class labels, with the goal of obtaining partitions that are better aligned with true classes \cite{basu2004probabilistic,basu2008constrained,liu2015clustering, cheng2007clustering,Law2007,lu2004semi}. One typical way of injecting class label information into clustering is in the form of pairwise \textit{constraints} (typically, \textit{must-link} and \textit{cannot-link} constraints), or pair-wise \textit{preferences} (\textit{e.g.}, \textit{should-link} and \textit{shouldn't-link}), which indicate whether a given pair of points is believed to belong to the same or different classes. 

Most SSC approaches rely on adapting existing unsupervised clustering methods to handle partial (namely, pairwise) information \cite{melnykov2016semi,bilenko2004integrating, cheng2007clustering,Law2007,lu2004semi,qian2016affinity}. This requires  transferring class-label knowledge into a clustering algorithm, which is often unnatural and puts a higher weight on clustering structure than on class labels. It has been recently shown that discriminative clustering methods, which approach clustering problems using classification tools, are usually more effective in taking advantage of label constraints/information \cite{pei2016comparing, smieja2018semi}. While those formulations assume that class labels are the primary source of semi-supervision, it might be difficult to produce satisfactory results in the presence of a large number of clusters. In that case, a small number of pairwise constraints may not allow to determine the correct clusters assignments.

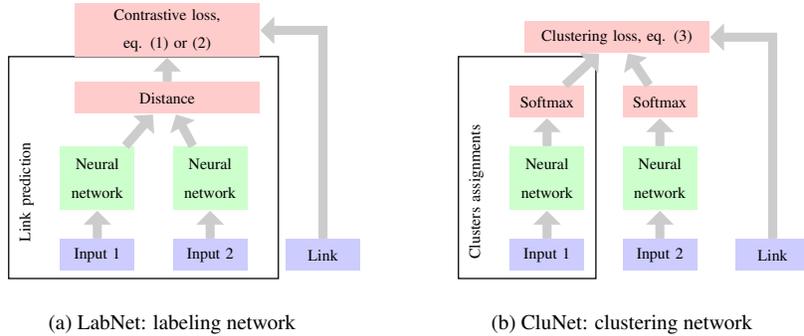
\begin{figure*}[t]
    \centering
    \subfloat[\SLN{}: labeling network\label{fig:siam}]{
        \scalebox{.6}{
\tikzstyle{data} = [rectangle, fill=blue!20, 
    text width=4em, text centered, minimum height=2em]
\tikzstyle{nn} = [rectangle, fill=green!20, 
    text width=4em, text centered, minimum height=4em]
\tikzstyle{fun} = [rectangle, fill=red!20, 
    text width=4em, text centered, minimum height=2em]
\tikzstyle{loss} = [rectangle, fill=red!20, 
    text width=11em, text centered, minimum height=2em]
\tikzstyle{back} = [rectangle, draw=black, 
    text width=8em, text centered, minimum height=14em, minimum width=17em]
\tikzstyle{back2} = [rectangle, draw=none, 
    text width=12em, text centered, rotate=90]

\tikzstyle{myarrows}=[line width=0.75mm,draw=gray!40,-triangle 90,postaction={draw, line width=2mm, shorten >=3mm, -}]

\begin{tikzpicture}[thick,node distance=2.5cm]
  \node[data] (in1) {Input 1};
  \node[back, above right=-0.9cm and -8em of in1] (back) {};
  \node[back2, above right=-1.2cm and -6em of in1] (back2) {Link prediction};
  \node[data,right of=in1] (in2) {Input 2};
  \node[data,right of=in2] (label) {Link};
  \node[nn,above of =in1, node distance=1.7cm] (nn1) {Neural network};
  \node[nn,above of=in2, node distance=1.7cm] (nn2) {Neural network};
  \node[loss,above right=0.7cm and -1.35cm of nn1] (dist) {Distance};
  \node[loss,above of=dist, node distance=1.5cm] (loss) {Contrastive loss, eq. \eqref{eq:sln} or \eqref{eq:sssln}};
  \draw[myarrows] (in1) -- (nn1);
  \draw[myarrows] (in2) to (nn2);
  \draw[myarrows] (nn1) to (dist);
  \draw[myarrows] (nn2) to (dist);
  \draw[myarrows] (dist) to (loss);
  \draw[myarrows] (label) |- (loss);
\end{tikzpicture}}
    }\qquad
    \subfloat[\CN{}: clustering network\label{fig:clus}]{
        \scalebox{.6}{
\tikzstyle{data} = [rectangle, fill=blue!20, 
    text width=4em, text centered, minimum height=2em]
\tikzstyle{nn} = [rectangle, fill=green!20, 
    text width=4em, text centered, minimum height=4em]
\tikzstyle{fun} = [rectangle, fill=red!20, 
    text width=4em, text centered, minimum height=2em]
\tikzstyle{loss} = [rectangle, fill=red!20, 
    text width=11em, text centered, minimum height=2em]
\tikzstyle{back} = [rectangle, draw=black, 
    text width=8em, text centered, minimum height=14em]
\tikzstyle{back2} = [rectangle, draw=none, 
    text width=12em, text centered, rotate=90]

\tikzstyle{myarrows}=[line width=0.75mm,draw=gray!40,-triangle 90,postaction={draw, line width=2mm, shorten >=3mm, -}]

\begin{tikzpicture}[thick,node distance=2.5cm]
  \node[data] (in1) {Input 1};
  \node[back, above right=-0.9cm and -8em of in1] (back) {};
  \node[back2, above right=-1.2cm and -6em of in1] (back2) {Clusters assignments};
  \node[data,right of=in1] (in2) {Input 2};
  \node[data,right of=in2] (label) {Link};
  \node[nn,above of =in1, node distance=1.7cm] (nn1) {Neural network};
  \node[nn,above of=in2, node distance=1.7cm] (nn2) {Neural network};
  \node[fun,above of=nn1, node distance=1.7cm] (f1) {Softmax};
  \node[fun,above of=nn2, node distance=1.7cm] (f2) {Softmax};
  \node[loss,above right=0.7cm and -1.35cm of f1] (loss) {Clustering loss, eq. \eqref{eq:loss}};
  \draw[myarrows] (in1) -- (nn1);
  \draw[myarrows] (in2) to (nn2);
  \draw[myarrows] (nn1) to (f1);
  \draw[myarrows] (nn2) to (f2);
  \draw[myarrows] (f1) to (loss);
  \draw[myarrows] (f2) to (loss);
  \draw[myarrows] (label) |- (loss);
\end{tikzpicture}}
    }
    \caption{Illustration of the proposed \our{} model. The labeling network is trained to label new pairs as must-link or cannot-link constraints. The clustering network is trained on the set of pairwise constraints generated by the labeling network to predict final clusters assignments.}
    \label{fig:intro}
\end{figure*}

In this paper, we go one step further than other discriminative approaches and decouple SSC into two stages:
\begin{description}
    \item[Stage 1:] predict pairwise relations between pairs of unlabelled points (a binary classification problem), which allows assigning predicted labels to unlabeled pairs, thus increasing the number of labeled pairs,
    \item[Stage 2:] use the labeled pairs (both given and predicted) in a semi-supervised clustering method.
\end{description}
The rationale behind our approach follows from the observation that it is easier to learn a binary classifier than to solve a multiclass problem under partial supervision, especially when the number of classes (clusters) is high. To increase the flexibility of our framework, we instantiate it with two neural networks, specifically with so-called Siamese neural networks \cite{Bromley,koch2015siamese}. The first network (\SLN--\textit{labeling network}) is used to classify pairs of examples as must-link or cannot-link constraints, while the second network (\CN--\textit{clustering network}) is trained on labeled pairs to predict final clusters assignments (see Figure \ref{fig:intro}). We term our method \our -- {\bf S}emi-{\bf S}upervised \textbf{S}iamese \textbf{C}lassifiers for \textbf{C}lustering.

In the experiments reported below, we implement \our{} with general-purpose dense \textit{deep neural networks} (DNNs) as well as with \textit{convolutional neural networks} (CNNs) to handle images. In both cases, \our{} outperforms other neural-network-based SSC techniques. Additionally, we experimentally and theoretically analyze the impact of the networks' parameterization on the clustering results. Our contributions are summarized as follows:

\begin{enumerate}
    \item a classification-based method for SSC with pairwise constraints, which first labels pairs of data points and then uses these predicted labels to perform SSC;
    \item an implementation of the proposed method with two Siamese DNNs, allowing to control the flexibility of the clustering model by adjusting the numbers of layers and neurons; the corresponding parameterization is studied theoretically and experimentally.
    \item experimental results showing the superiority of the proposed \our{} method over related approaches on several datasets, including \textit{Letters}, which has 26 classes; to the best of our knowledge, SSC had not been tested on such a large number of classes.
\end{enumerate}

The code of our method will be made publicly available online after acceptance of the paper.

\section{Related work}
The most common way of using pairwise constraints in SSC relies on modifying the underlying cost function of a classical unsupervised clustering models \cite{smieja2017constrained, qian2016affinity, lu2016semi}. Such an approach was used in k-means, using a term penalizing pairwise constraint violation \cite{bilenko2004integrating}, and in \textit{Gaussian mixture models} (GMM), with hidden Markov random fields modelling pairwise relations \cite{basu2004probabilistic,  cheng2007clustering,  lu2004semi}. In spectral clustering, the underlying eigenvalue problem was modified by adding the pairwise constraints to the corresponding objective function \cite{kawale2013constrained,Wang_FCSC}. Another line of work focuses on modifying the similarity measure based on the pairwise relations \cite{Asafi_ConstraintsFeatures, chang2014learning, wang2012constraint}, by learning optimal Mahalanobis distances \cite{davis2007information,xing2003distance}, or more general kernel functions \cite{yin2010semi}. 

Recently, it has been shown that discriminative clustering formulations  \cite{Kaski2005} are often more effective in leveraging pairwise relations than the aforementioned methods. The authors of \cite{pei2016comparing} used an analogue of the classification log-loss function based on pairwise constraints and added entropy regularization \cite{krause2010discriminative} to prevent degenerate solutions. In a similar spirit, \cite{smieja2018semi} maximized the expected number of correctly classified pairs based on pairwise constraints and an underlying distance function. The authors of \cite{calandriello2014semi} used a squared-loss mutual information to regularize a discriminative clustering model.

Although DNNs are dominant in many areas of machine learning, their have rarely been  used for SSC. The authors of \cite{hsu2015neural} used a KL-divergence-based loss to train a DNN to predict cluster distribution from pairwise relations; one limitation of that method is its inability to use unlabeled data. Other works \cite{fogel2019clustering, shukla2018semi, zhang2019deep} used auto-encoders with reconstruction losses to exploit inner characteristics of unlabeled data. In \cite{shukla2018semi}, the k-means loss is combined with KL-divergence to create compact clusters preserving pairwise relations. In \cite{fogel2019clustering}, the distance between must/cannot-link pairs was minimized/maximized, instead of using KL-divergence. \textit{Deep embedding clustering} (DEC) is a method that jointly learns feature representations and cluster assignments using deep neural networks \cite{xie2016unsupervised}. Finally, a method capable of using various types of side information has been proposed in \cite{zhang2019deep}.

Our work extends recent discriminative SCC methods \cite{smieja2018semi, pei2016comparing} by learning additional pairwise relations. Moreover, the approach is implemented using Siamese neural networks \cite{Bromley,hadsell2006dimensionality, koch2015siamese}, allowing for higher flexibility. In contrast to the aforementioned deep SSC methods, our model is fully discriminative and uses misclassification error as the only loss term. 

\section{Proposed Method}
\subsection{Formulation}
\label{sec:formulation}

Let $\X \subset \R^D$ be a dataset, where every instance $x \in \X$ belongs to one of $K$ classes. The goal is to split $\X$ into $K$ clusters, which are compatible with the true (unknown) classes.

We assume that partial class information is given in the form of pairwise constraints, indicating whether two examples belong to the same (\textit{must-link} constraint) or different (\textit{cannot-link } constraint) classes. Formally, the class information is expressed via a set $\LL \subset \X \times \X$, where $\LL = \M \cup \C$, and 
\begin{align*}
\M=\{(x,y) \in \X \times \X\! : x \text{ and } y \; \text{belong to the same class}\},\\
\C=\{(x,y) \in \X \times \X\! : x \text{ and } y \; \text{belong to different classes}\}.
\end{align*}
To make the notation lighter in what follows, we assume that $\M$ always contains all the $|\X|$ pairs of the form $(x,x)$, because the binary relation ``belong to the same class" is obviously reflexive. Furthermore, because the binary relations ``belong to the same class" and ``belong to different classes" are both symmetric, we also assume that
\[
(x,y) \in \M \Rightarrow (y,x) \in \M  \;\;\; \mbox{and} \;\;\; (x,y) \in \C \Rightarrow (y,x) \in \C.
\]
Finally, let $\U = \X^2\setminus \LL\, $ denote the set of unlabeled pairs.

The proposed \our{} model is composed of two classification neural networks. The labeling network (\SLN)  is trained to assign labels (must-link or cannot-link) to new pairs of examples not in $\LL$. The clustering network (\CN)  is trained to use labeled pairs to predict clusters assignments. The proposed scheme is illustrated in Figure~\ref{fig:intro} and next described in detail. 

\subsection{The Labeling Network: \SLN}
\label{sec:lab}
Instead of doing SSC directly, which can be a difficult multi-class problem, we first tackle a simpler binary classification problem: learning to label new pairs (\textit{i.e.}, not in $\LL$) as belonging to the same class (must-link) or different classes  (cannot-link). By classifying instances in $\U$ but not in $\LL$, we obtain new must-link/cannot-link labels that will be used by \CN\ to predict the final clusters assignments (as described in the next subsection).


We address this classification problem using a pair of Siamese neural networks (identical networks, \textit{i.e.}, with shared weights) \cite{koch2015siamese}. The task of these networks is to take a pair of points $(x,y)$ and return their representations $\bigl( h(x),h(y)\bigr)$, based on which it will be decided if $x$ and $y$ are in the same or different classes. Naturally, they are trained to make $h(x)$ \textit{close} to $h(y)$, if $(x,y) \in \M$, and $h(x)$ \textit{distant} from $h(y)$, if $(x,y) \in \C$. To this end, we use a \textit{contrastive} loss based on the Euclidean distance $d(x,y) = \|h(x) - h(y)\|$, defined as:
\begin{equation} \label{eq:sln}
\sln(\M,\C) =
\frac{1}{|\LL|} \biggl( \sum_{(x,y) \in \M} d(x,y)^2 \\  + \sum_{(x,y) \in \C} \max\{1-d(x,y),0\}^2 \biggr).
\end{equation}
Notice that the presence of pairs of the form $(x,x)$ in $\M$ does not contribute to $\sln(\LL)$ because $d(x,x) = 0.$

\begin{figure}[t]
    \centering
    \includegraphics[width=0.5\textwidth]{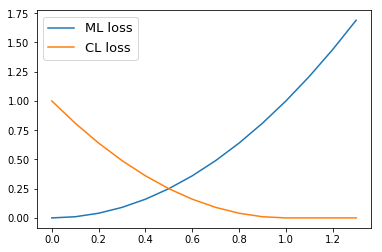}
    \caption{Contrastive loss of labeling network for must-link and cannot-link pairs.}
    \label{fig:contrastive}
\end{figure}

Clearly, being a distance, $d(x,y) \geq 0$, for all $x,y\in\X$. Observe that a cannot-link pair contributes to the loss only if its distance is below 1, see Figure \ref{fig:contrastive}. A crucial aspect is that \SLN\  does not decide whether two points belong to the same or different classes; it only yields similarity scores for pairs of data points\footnote{Siamese networks have been used for one-shot learning \cite{koch2015siamese}, where the class of a given example is decided by comparing the output one of the twin networks with that of the other on a set of examples of known classes.}. A hard link prediction is obtained by comparing the distance $d(x,y)$ with a threshold $T$: $x$ and $y$ are classified as being in the same class if and only if $d(x,y)^2 < T$. One natural choice is $T = 1/2$, because if $d(x,y)^2 = 1/2$, then $(1-z(x,y))^2 = 1/2$ as well. Below, we will explain that $T < 1/2$ is usually a better choice in our case.

In the training phase of the \SLN{}, only pairwise constraints are used (the loss in \eqref{eq:sln} only depends on ${\cal L}$). To leverage information contained in unlabeled data, we consider an adaptation of SEVEN (\textit{SEmi-supervised VErification Network} \cite{noroozi2017seven}), yielding a  semi-supervised version of \SLN. The idea is to encourage the mapping $h$ to learn a salient structure shared by all categories. For this purpose, each Siamese twin is supplied with a decoder network $g$, which aims at obtaining a reconstruction of $x$ from its latent representation $h(x)$: $\hat{x} = g(h(x))$. This goal is pursued by using a \textit{reconstruction} error loss term,
$$
\mse(\X) = \frac{1}{|\X|^2} \sum_{(x,y) \in \X\times\X} \bigl( \|\hat{x} - x\|^2 + \|\hat{y} - y\|^2 \bigr).
$$
Finally, the total loss used for training the semi-supervised \SLN{} is
\begin{equation} \label{eq:sssln}
 \sln(\M,\C) + \lambda \; \mse(\X),
\end{equation}
where $\lambda$ is a trade-off parameter.

Once trained, the \SLN{} is applied to yield pairwise constraints for all pairs of data points. Let 
\begin{eqnarray*}
\tilde\M & = & \M \cup \{(x,y) \in \U : d(x,y)^2 < T \} \\
\tilde\C & = & \C \cup \{(x,y) \in \U : d(x,y)^2 \geq T \};
\end{eqnarray*}
clearly, $\tilde\M \cup \tilde\C = \X^2$.

\subsection{The Clustering Network: \CN}
\label{sec:CluNet}

Since, by the application of the \SLN\ to the unlabeled pairs yields pairwise constraints to all the pairs in the dataset, the final clustering can be obtained in a purely supervised manner. Instead of a typical unsupervised clustering method (\textit{e.g.}, k-means or GMM), we thus employ a discriminative framework, which is more effective in the supervised case. Namely, we directly model cluster assignments with posterior probabilities $p_k(x) = p(k|x)$, for $k=1,\ldots,K$. From these posterior probability  estimates, $\X$ may be partitioned by assigning every point $x\in\X$ to the cluster $k$ that maximizes $p_k(x)$.


To provide sufficient flexibility, we instantiate the \CN\ as a Siamese pair of identical DNNs, where each pair of points $(x,y)$ is processed by two identical (Siamese twins) sub-networks with shared weights. Equipped with \textit{softmax} output layers, these Siamese twin networks yield class posterior probabilities $p_1(x), \ldots,p_K(x)$ and $p_1(y), \ldots,p_K(y)$, for each pair of items $(x,y)$ .


To form clusters consistent with pairwise constraints, we aim at minimizing the number of misclassified pairs. Note that, given  the posterior class probabilities $p_1(x), \ldots,p_K(x)$ and $p_1(y), \ldots,p_K(y)$, for a pair of points $(x,y)$,  the probability that $x$ and $y$ are in the \textit{\textbf{s}ame} cluster is given by
$$
p^{s}(x,y) = \sum_{k=1}^K p_k(x) \; p_k(y)
$$
whereas $p^{d}(x,y) =1 - p^{s}(x,y)$ is the probability that they are in \textit{\textbf{d}ifferent} clusters. We thus define the misclassification loss with respect to the must-link and cannot-link information as
\begin{equation} \label{eq:loss}
\cn(\tilde\M,\tilde\C) = \frac{1}{|\X|^2} \Bigl( \sum_{(x,y) \in \tilde\M} \!\!\! (1-p^{s}(x,y)) \; + \!\!\! \sum_{(x,y) \in \tilde\C} \!\! p^{s}(x,y) \Bigr).
\end{equation}

The structure of the \CN{} is shown in Figure \ref{fig:clus}. Whereas during the training phase, the loss function in \eqref{eq:loss} uses the Siamese pair since it applies to pairs of points, in the testing phase, only one of the networks is needed (as indicated in  Figure \ref{fig:clus}) to produce cluster assignments, where a given point $z\in\mathbb{R}^D$ is assigned to the cluster with the highest posterior probability: $\hat{k}(x) = \arg\max_k p_k(z)$.

\subsection{Adjusting the \SLN\  Classification Threshold} \label{sec:th}
We analyze the influence of the \SLN{} threshold $T$ on the \CN{} results. Let us begin by assuming that $T = 0$; in this case, all pairs in $\U$ are labeled by the \SLN\ as cannot-links (\textit{i.e.}, $\tilde\C = \C\cup\U$ and $\tilde\M = \M$) and the loss in \eqref{eq:loss} can be written as 
\[
\cn(\tilde\M,\tilde\C) = \frac{1}{|\X|^2} \Bigl( 
\sum_{(x,y) \in \X^2} \!\!\! p^s (x, y) - 2 \!\!\! \sum_{(x,y) \in \M} \!\! p^s (x, y) + |\M|\Bigl).
\]
Assuming $|\M| \ll \X^2$ (as will be the case in all the experiments below and is the typical scenario in SSC),
\begin{eqnarray}
\cn(\tilde\M,\tilde\C) & \simeq &  \frac{1}{|\X|^2} \sum_{(x,y) \in \X^2} p^s (x, y)\nonumber \\ & = & \frac{1}{|\X|^2} \sum_{(x,y) \in \X^2} \sum_{k=1}^K p_k(x) p_k(y) \nonumber\\
 & = & \sum_{k=1}^K \left(\frac{1}{|X|} \sum_{x \in X}  p_k(x)\right)^2 
 = \sum_{k=1}^K \hat{p}_k^2, \label{eq:tsallis}
\end{eqnarray}
where 
\[
\hat{p}_k = \frac{1}{|\X|} \sum_{x \in \X}  p(k|x)  
\]
is an estimate of the probability of $k$-th class approximated by the clustering model (notice that $\sum_{k=1}^K \hat{p}_k = 1$). The last term in \eqref{eq:tsallis} is related to the index-2 Tsallis entropy $S_2(p_1,\ldots,p_K)$ \cite{furuichi2006information}
$$
\sum_{k=1}^K \hat{p}_k^2 = 1 - S_2(\hat{p}_1,\ldots,\hat{p}_k).
$$
Since $S_2$ is maximized by the uniform distribution, $\cn(\tilde\M,\tilde\C)$ is (approximately) minimized by taking equally-sized clusters. This means that by predicting a large number of cannot-link pairs (by setting $T=0$) encourages high entropy (approximately uniform) clusterings and discourages degenerate solutions.


By increasing the threshold $T$, more pairs are classified as must-link and fewer as cannot-link. In this case, 
\[
\cn(\tilde\M,\tilde\C) = \frac{1}{|\X|^2} \Bigl( 
\sum_{(x,y) \in \X^2} \!\!\! p^s (x, y) - 2 \!\!\! \sum_{(x,y) \in \tilde\M} \!\! p^s (x, y) + |\tilde\M|\Bigl),
\]
where (above a certain value) $|\tilde\M| > |\M|$. This can be rewritten as
\[
\cn(\tilde\M,\tilde\C) =\sum_{k=1}^K \hat{p}_k^2 - \frac{2}{|\X|^2} \sum_{(x,y) \in \tilde\M} \!\! p^s (x, y) + \frac{|\tilde\M|}{|\X|^2},
\]
where the last term is a constant that depends only on the output of the \SLN.
This form of the loss function shows that: (a) it encourages pairs in $\tilde\M$ to be given high probability of being classified in the same class (large $p^s (x, y)$); (b) it encourages the Tsallis entropy of the estimated class probabilities to be high (low $\sum_{k=1}^K \hat{p}_k^2$). In other words, must-link constraints (those in $\tilde\M$) play a more active role in this loss function, whereas cannot-link pairs (those in $\tilde\C$) essentially only contribute to the entropic term of the loss.

The observation in the previous paragraph shows that obtaining must-link constraints is crucial for the performance of the \CN{}. This is however a double-edged sword; correct must-links provide valuable information to train the \CN{}, but erroneous ones may be very harmful. If two instances from different classes are wrongly put in $\tilde\M$, this directly impacts the middle term of the loss $\cn$, whereas two examples from the same class that are wrongly put in $\tilde\C$ essentially only affect the regularization term (first term of $\cn$), in addition to being missing from $\tilde\M$.
Furthermore, erroneous must-link constraint can be implicitly propagated to other pairs due to the transitivity of the binary relation ``belong to the same class", whereas the binary relation ``belong to different classes" is not transitive.

The above considerations suggest that it is safer to use small values of threshold $T$. This is especially important if the number of given pairwise constraints is small, because the accuracy of \SLN{} may then be low. In this case, the \SLN{} with a small $T$ puts in $\tilde\M$ only pairs about which it is very confident. The other pairs will contribute to the entropic regularization term. If the number of given constraints is larger, we can use a higher threshold $T$ and label more pairs as must-link with higher confidence. Consequently, $T = 1/2$ may be optimal only in the presence of large sets of constrains, which is seldom the case in practice. Experimental validation of this rationale is presented in Section \ref{sec:thr}.

\begin{figure*}[h]
    \centering
    \includegraphics[width=0.48\textwidth]{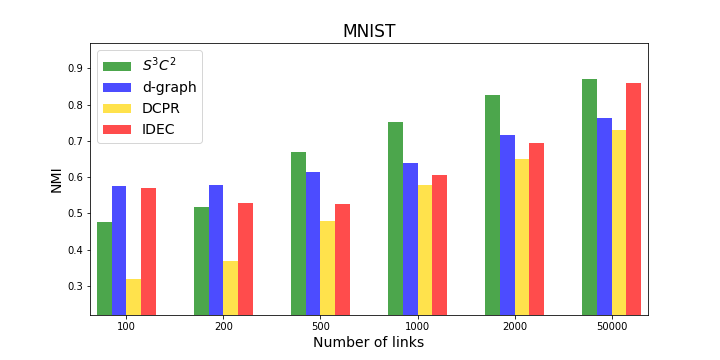} \includegraphics[width=0.48\textwidth]{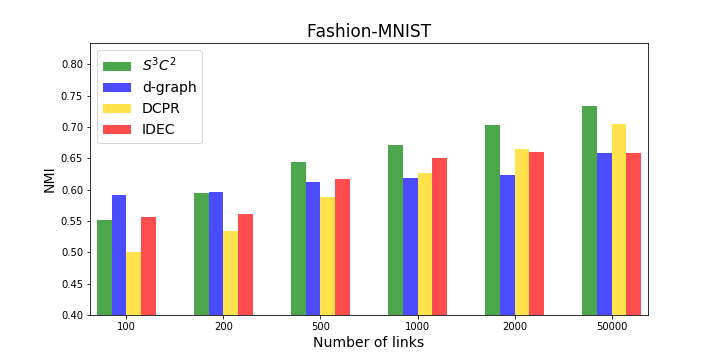}
    \includegraphics[width=0.48\textwidth]{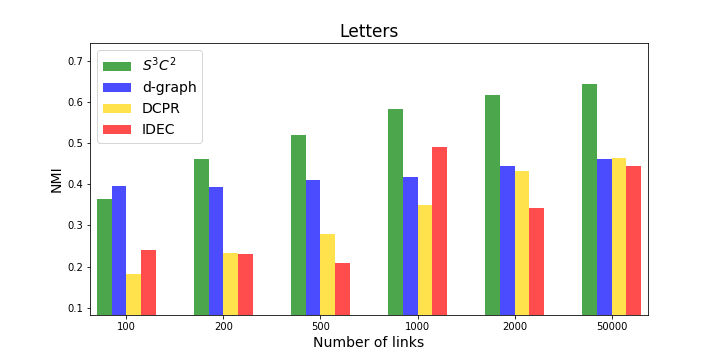}
    \includegraphics[width=0.48\textwidth]{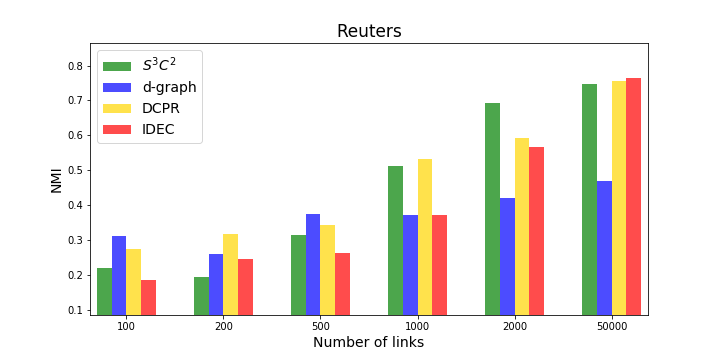}
    \caption{Performance of clustering models on four data types with varied numbers of constraints.}
    \label{fig:res1}
\end{figure*}

\begin{figure*}[h]
    \centering
    \includegraphics[width=0.48\textwidth]{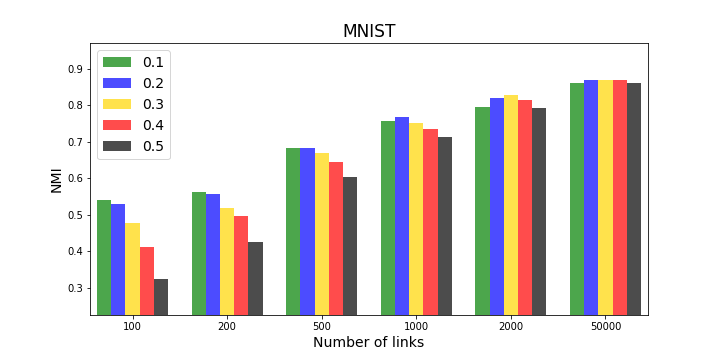} \includegraphics[width=0.48\textwidth]{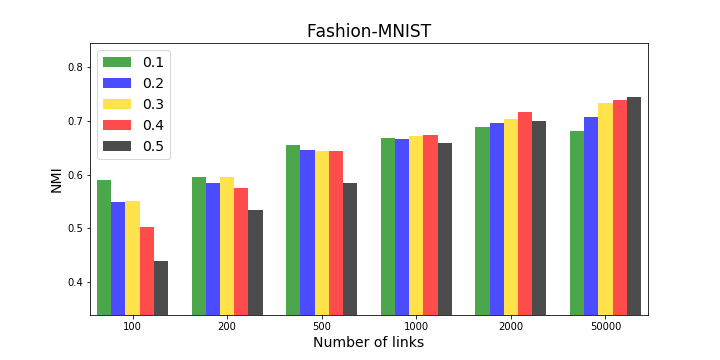}
    \includegraphics[width=0.48\textwidth]{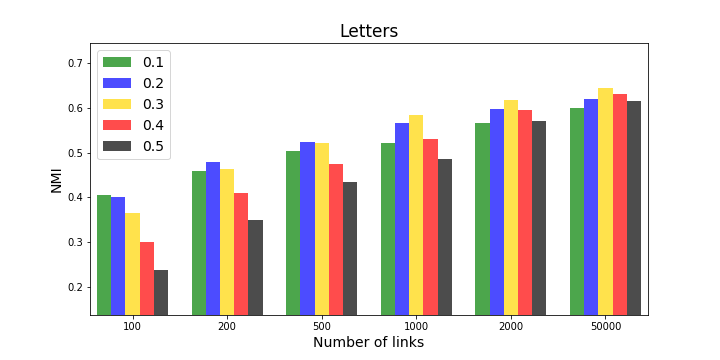}
    \includegraphics[width=0.48\textwidth]{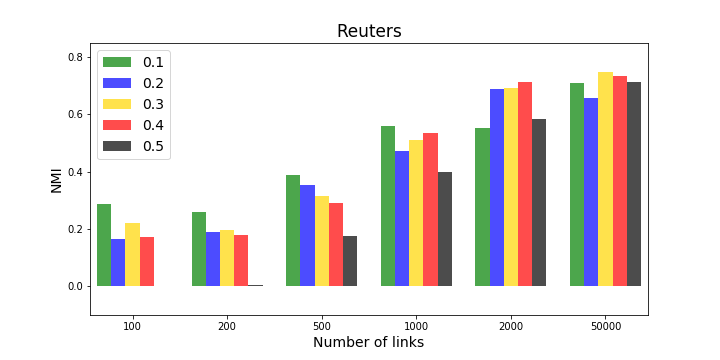}
    \caption{Performance of \our{} with different threshold $T$.}
    \label{fig:res2}
\end{figure*}

\section{Experiments}
In this section, we evaluate our approach \our{} against state-of-the-art methods and investigate the effect of the parametrization of the \SLN{} on the clustering results.

\begin{figure*}[h]
    \centering
    \includegraphics[width=0.48\textwidth]{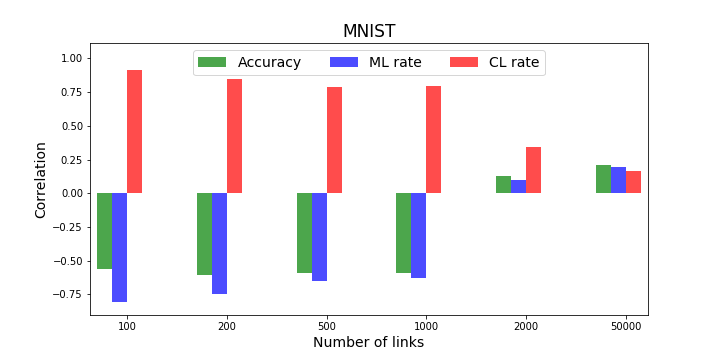} \includegraphics[width=0.48\textwidth]{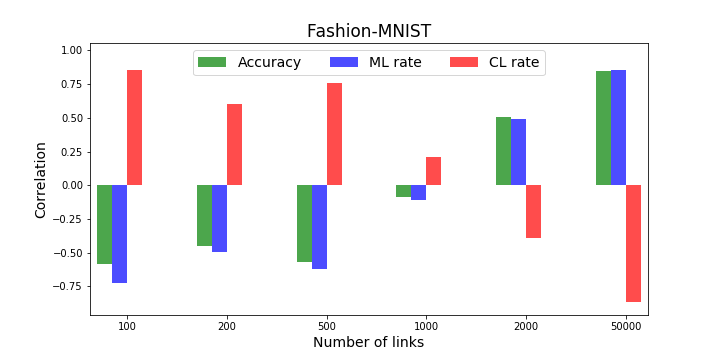}
    \includegraphics[width=0.48\textwidth]{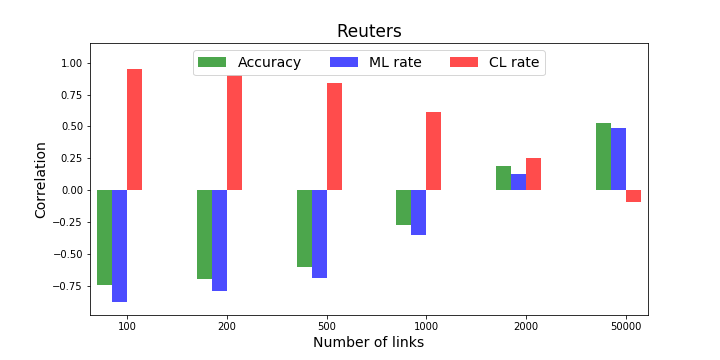}
    \includegraphics[width=0.48\textwidth]{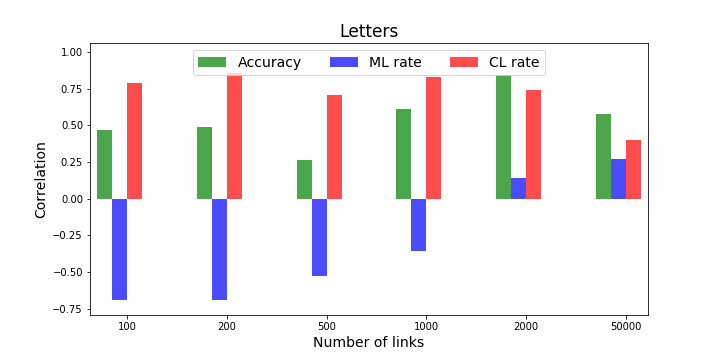}
    \caption{Correlation between clustering NMI and three indicators of \SLN{}: accuracy, ML rate, and CL rate.}
    \label{fig:corr}
\end{figure*}

\subsection{Experimental setting}

We consider four popular datasets with normalized attributes: 
\begin{itemize}
    \item {\bf MNIST:} It contains 70k gray scale images of handwritten digits of the size $28 \times 28$ (60k for training and 10k for testing) \cite{lecun1998gradient}. The set is divided into 10 equally-sized classes.
    \item {\bf Fashion-MNIST:} It is a dataset of 70k gray scale images of the size $28 \times 28$ (60k for training and 10k for testing) with 10 classes \cite{xiao2017fashion}. Images show clothing items. 
    \item {\bf Reuters:} This dataset contains English news stories labeled with a category tree \cite{lewis2004rcv1}. Analogically to previous uses of this data in clustering, we randomly sampled a subset of 12k examples (10k for training and 2k for testing) from 4 root classes: corporate/industrial, government/social, markets and economics. Documents were represented using TF-IDF features on the 2000 most frequent words.
    \item {\bf Letters:} This dataset contains a description of capital letters in the English alphabet (26 classes) \cite{frey1991letter}. The character images were based on 20 different fonts and each letter within these 20 fonts was randomly distorted to produce 20k examples. Each example was converted into 16 primitive numerical attributes (statistical moments and edge counts). We used 15k first examples for training and remaining 5k for testing.
\end{itemize}

To generate pairwise constraints, we randomly select $l \in \{100, 200, 500, 1000, 2000, 5000\}$ pairs of instances and label them either as must-link or cannot-link constraints (depending on their true relations). The number of must-links and cannot-links are kept equal. The results are evaluated using \textit{normalized mutual information} (NMI \cite{Strehl_NMI_2002}), which attains a maximal value 1 for two identical partitions. To reduce the effect of randomness, we generate 5 different sets of pairwise constraints for each number of constraints $l$; the final score is the NMI average over these 5 sets.

\subsection{Comparison with related models}
We first compare the performance of \our{} with other SSC approaches for various levels of pairwise constraints. We restrict our attention to the DNN-based methods.

As explained in Section \ref{sec:CluNet}, each of the networks in the Siamese pair in \CN{} is equipped with a softmax output layer. To make our model domain-agnostic, rather than specialized to a specific dataset or domain, we use two dense hidden layers with 256 neurons each and ReLU activation function, as well as dropout after each hidden layer (with rate 0.1, except for the Reuters dataset where we use dropout rate of 0.5). Each batch consists of 100 training pairwise constraints and 1000 unlabeled pairs labeled by the \SLN{}. The learning rate is set to $10^{-3}$. The \SLN{} has an analogous structure: each DNN in Figure \ref{fig:intro}(a), which corresponds to the mapping $h$ introduced in Section \ref{sec:lab}, has 2 hidden dense layers with 256 neurons each, ReLU activation function, and dropout with the same rates, and an output dense layer also with 256 neurons, but with sigmoid activation function (\textit{i.e.}, $h(x)\in [0,1]^{256}$). The threshold $T$ in $\sln$ is set to $0.3$. A more detailed study of the selection of $T$ is presented below. We use batch size of 256 examples and learning rate of $10^{-3}$. We restrict our attention to fully supervised version of \SLN{}.

For comparison we select three recent SSC methods:
\begin{itemize}
    \item {\bf d-graph:} this is a DNN-based implementation of d-graph \cite{smieja2018semi}. The network architecture is identical to \CN{} (the batch structure is also the same). The closest $30$ unlabeled pairs in each batch are labeled as auxiliary must-link constraints, while the remaining pairs are considered as cannot-link\footnote{We also tried different numbers of neighbors, but the results were worse.}.  
    \item {\bf DCPR:} this is a DNN-based implementation of DCPR \cite{pei2016comparing} (the architecture and the structure of batch is the same as in d-graph). The entropy and conditional entropy used to regularize the clustering model are estimated from each batch.
    \item {\bf IDEC:} this is a SSC method proposed in \cite{zhang2019deep} (\url{http://github.com/blueocean92}) using pairwise constraints. The network structure and training procedure follow the author's code.
\end{itemize}

The results presented in Figure \ref{fig:res1} show the good performance of \our{}, specially with the larger numbers of constraints. For the smaller numbers of constraints (100 and 200 links), \SLN{} is not able to accurately predict links,  negatively influencing the performance of \CN. In this case, \our{} is inferior to d-graph, but it is still competitive with or better than DCPR and IDEC. It is worth  emphasizing the extremely good results on the Letters dataset, which is composed of 26 classes. To the best of our knowledge, a dataset with so many classes had not been used before for SSC with pairwise constraints\footnote{A subset of the Letters dataset with only 5 classes was used in \cite{smieja2018semi,pei2016comparing}.}.

The results in Figure \ref{fig:res1} show that d-graph performs best with the smaller number of pairwise constraints. For higher number of constraints, it is outperformed by \our\ and IDEC. This is arguably due to the fact that d-graph generates auxiliary labeled pairs based only on distances. Moreover, a single batch may be too small to find a good k-NN graph. DCPR is competitive with \our{} only on the Reuters dataset, but for other datasets its performance is worse. IDEC gives good results for large number of constraints, but its performance is not stable (its results do not always increase as the number of constraints grows for MNIST and Letters).

\begin{figure*}[h]
    \centering
    \includegraphics[width=0.48\textwidth]{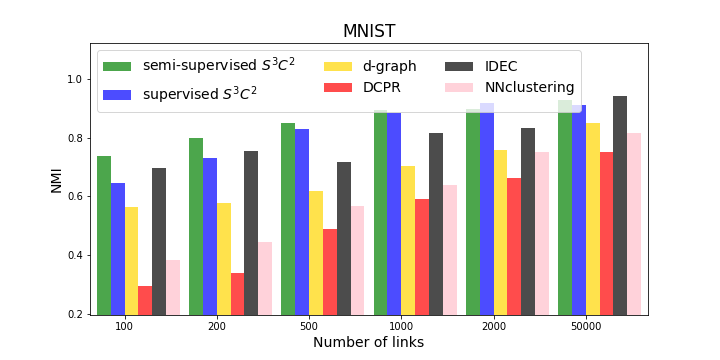} \includegraphics[width=0.48\textwidth]{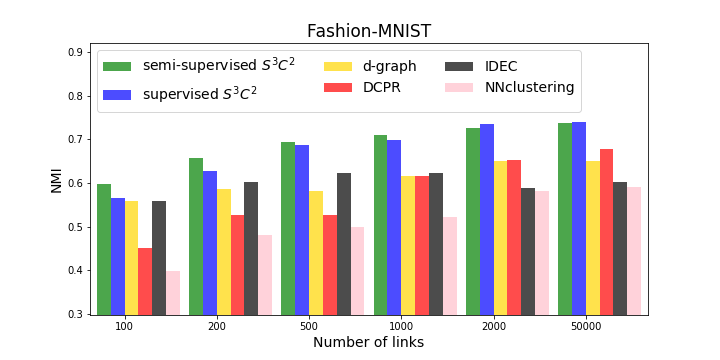}
    \caption{Performance of clustering methods with convolutional architecture applied to image data.}
    \label{fig:images}
\end{figure*}

\begin{figure*}[h]
    \centering
    \includegraphics[width=0.48\textwidth]{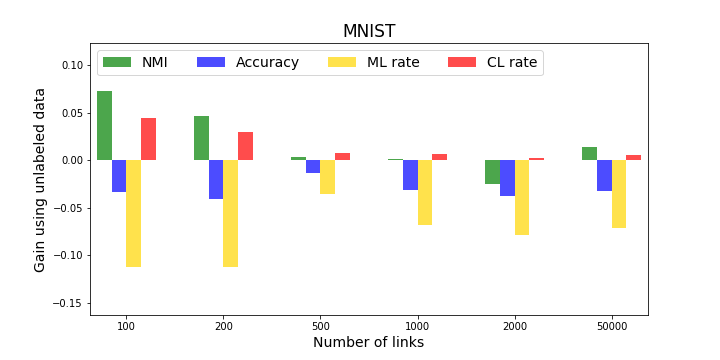} \includegraphics[width=0.48\textwidth]{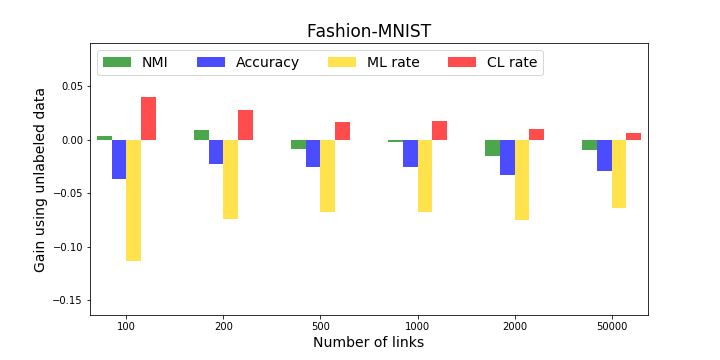}
    \caption{Difference between learning statistics of \our{} using semi-supervised and supervised \SLN{} (clustering NMI as well as accuracy, ML rate, and CL rate of labeling networks).}
    \label{fig:images2}
\end{figure*}

\subsection{Study of the Labeling Network} \label{sec:thr}
As discussed in Section \ref{sec:th}, the choice of the labeling threshold $T$ may be crucial for performance of \our{}. Since it may be difficult to find an optimal value using cross-validation, if only a small number of labeled pairs is available, we experimentally analyze various threshold values to get better insight into our model.

The results presented in Figure \ref{fig:res2} are consistent with the reasoning presented in Section \ref{sec:th}. For small numbers of given constraints (small $|\LL|$), \SLN{} is unable to correctly predict pairwise relations. It is thus better to use a low thresholds and assign must-link constraints only to the most confident pairs, because erroneous must-link constraints negatively affect the clustering results and, as argued in Section \ref{sec:th}, cannot-link constraints have essentially a regularization effect. For larger numbers of labeled pairs, a higher threshold can be used due to the better accuracy of \SLN{}. Nevertheless, it is difficult to define a general rule for threshold selection, but it can be seen that $T \in [0.2, 0.3]$ is a safe choice leading to good results for all datasets at all levels of semi-supervision.

To get further insight into our model, we compute the correlation coefficients between the clustering NMI and the classification statistics gathered from the \SLN{}. Namely, we consider: (a) accuracy; (b) must-link (ML) rate, $r(\M)$; (c) cannot-link (CL) rate, $r(\C)$. These quantities are defined as
$$
\begin{array}{l}
r(\M) = \frac{1}{|\M|} \sum\limits_{(x,y)\in \M} \1_{d(x,y)^2 < T},\\[1.8ex]
r(\C) = \frac{1}{|\C|} \sum\limits_{(x,y)\in \C} \1_{d(x,y)^2 \geq T}.
\end{array}
$$
While accuracy measures the overall performance of \SLN{} classifier, ML and CL rates assess how the model predicts examples from underlying classes. Figure \ref{fig:corr} shows that for small and medium numbers of constraints, the CL rate has the highest correlation with the clustering performance as measured by NMI. It is also interesting to observe that, in most cases, the ML rate has negative correlation with NMI, which partially confirms our intuition that labeling cannot-link pairs as must-link has a negative effect on the final performance. On the other hand, assigning cannot-link labels to must-link pairs does not have a negative influence, because it simply leads to stronger regularization. Such a labelling does not improve the performance of clustering model, but it also does not deteriorate it. For the highest numbers of constraints (2k and 5k) the correlation with CL rate is not so strong (it is negative for Fashion-MNIST). We verified that, in that cases, CL rates were higher than 95\% for most models. Consequently, the clustering results could be only improved by increasing the ML rate. It is evident that the accuracy of \SLN{} cannot be used as the only indicator of final success. Clearly, higher accuracy allows obtaining better clustering results, but ML and CL rates give us more detailed information. In particular, it is important to use a labeling network which has high CL rate and only then one should care about ML rate.

\subsection{Model specialized to image processing}
In the previous experiments, we used dense neural networks, which can be applied to generic (not too high-dimensional) datasets regardless on their domain. We now show that the performance of our method can be further increased by selecting network architecture specialized to a given task. In particular, we present its specialization to image data, using the  MNIST and Fashion-MNIST datasets. In addition, we also consider semi-supervised version of \SLN{} \cite{noroozi2017seven}, which is trained on unlabeled pairs as well.

The \CN{} is instantiated using two convolutional layers (32 filters each) with max pooling and dropout after each one. This is followed by two dense layers (with 128 and 10 neurons, respectively) and dropout between them. The architecture of the \SLN{} is composed of identical convolutional layers with max pooling and dropout, followed by a single dense layer with 128 neurons. In the case of semi-supervised \SLN{}, every Siamese twin is supplied with a decoder network, which is implemented using symmetric deconvolution layers and upsampling. Based on the results presented in \cite{noroozi2017seven}, we use $\lambda = 0.05$ as a trade-off parameter in \eqref{eq:sssln}. The other models, d-graph, DCPR and IDEC, are implemented using analogous architectures. Additionally, we use NNclustering \cite{hsu2015neural}. In contrast to the other methods herein considered, NNclustering is trained only on the set of pairwise constraints (no unlabeled pairs are used); we use authors' code, where the method is implemented using convolutional LeNet networks. 

The results presented in Figure \ref{fig:images} demonstrate that specialized convolutional architecture allows to obtain better clustering results than using dense networks (see Figure \ref{fig:res1} for a comparison). Moreover, the use of unlabeled data in \SLN{} has positive influence on the final results. Our clustering method with semi-supervised \SLN{} noticeably outperforms its variant with supervised \SLN{} when a small number of constraints is available. For larger numbers of constraints, the difference is smaller, because the network has enough data to be trained. As before, both variants of our method are better than d-graph, DCPR, and NNclustering. IDEC is also inferior to our method except the case of 5000 constraints for MNIST, where it obtains the highest performance. 

In addition to clustering NMI, we also assessed accuracy, as well as the ML and CL rates for both versions of \SLN{}. The differences between these quantities for the semi-supervised and supervised \SLN{} are shown in Figure \ref{fig:images2}. The results demonstrate that semi-supervised \SLN{} yields higher CL rates if smaller number of constraints are given. This again confirms that our clustering method is very sensitive to erroneous ML constraints and a good labeling network should correctly predict most of cannot-link pairs. 

\section{Conclusion}
In this paper, we introduced a classification-based approach to semi-supervised clustering with pairwise constraints. It was shown that decomposing a semi-supervised clustering task into two simpler problems, classifying pairwise relations and then performing supervised clustering, is a better option than directly solving the original task. Our framework is implemented using of two Siamese neural networks and is experimentally shown to achieve state-of-the-art performance on several benchmark datasets. 

In the future, we plan to investigate different approaches for classifying pairwise relations. On the one hand, it is beneficial to construct a model that guarantees high CL rate. On the other hand, one could also design active learning mechanisms, which query must-link pairs with high probability, in order to strengthen the clustering model.

\section*{Acknowledgement} This work was carried out when M. \'Smieja was a Post-Doctoral Scholar at Instituto Superior T\'ecnico, University of Lisbon. The work was partially supported by the National Science Centre (Poland), grant no. 2016/21/D/ST6/00980.

\bibliography{ecai}

\end{document}